# A Spiking Neuron Synaptic Plasticity Model Optimized for Unsupervised Learning

**Mikhail Kiselev**
Chuvash State University
Cheboxary, Russia
`mkiselev@chuvsu.ru`

November 2, 2021

## Abstract

Spiking neural networks (SNN) are considered as a perspective basis for performing all kinds of learning tasks – unsupervised, supervised and reinforcement learning. Learning in SNN is implemented through synaptic plasticity – the rules which determine dynamics of synaptic weights depending usually on activity of the pre- and post-synaptic neurons. Diversity of various learning regimes assumes that different forms of synaptic plasticity may be most efficient for, for example, unsupervised and supervised learning, as it is observed in living neurons demonstrating many kinds of deviations from the basic spike timing dependent plasticity (STDP) model. In the present paper, we formulate specific requirements to plasticity rules imposed by unsupervised learning problems and construct a novel plasticity model generalizing STDP and satisfying these requirements. This plasticity model serves as main logical component of the novel supervised learning algorithm called SCoBUL (Spike Correlation Based Unsupervised Learning) proposed in this work. We also present the results of computer simulation experiments confirming efficiency of these synaptic plasticity rules and the algorithm SCoBUL.

## 1   Introduction and motivation

At present, the sound expectations exist that spiking neural networks (SNN) used as a basis for creation of new generation of artificial intelligence systems may extend significantly their functionality and application area comparatively to intelligent systems based on traditional neural networks. It has been proven experimentally that hardware implementations of SNN consume several orders of magnitude less energy than traditional neural network on comparable tasks [1]. Asynchronous nature of SNN operation leads to potentially unlimited scalability of SNN-based systems [2] as it is demonstrated on the modern neurocomputers like SpiNNaker [3]. Non-trivial involvement of signal transmission delays in information processing by SNN makes this type of neural networks suitable for processing dynamic data streams [4]. However, three is still a number of unsolved scientific problems in the field of SNN which hinder wide application of SNN to practical problems. Probably, the hardest of them is the problem of SNN learning.

Learning algorithms of traditional neural networks are based on the fact that they can be represented as smooth multi-dimensional functions. Their output values depend smoothly on the input values as well as on their synaptic weights. It makes possible to use gradient descent methods for their training treated as optimization of their synaptic weights. The error backpropagation algorithm based on gradient descent is a well-known platform for majority of approaches to learning of traditional neural networks. In contrast,

SNNs are discrete systems by their nature. The "all-or-nothing" behavior of spiking neurons makes direct application of gradient methods impossible. This problem can be overcome to some extent with help of the various "surrogate gradient" methods [5] but this solution is considered by many researchers as partial and not exploiting the natural advantages of SNNs. Seemingly, the more adequate approach to SNN learning is based on reproduction of the synaptic plasticity principles observed in living neuronal ensembles, the principles that utilize the basic concepts of SNN – asynchronous operation of neurons and spiking nature of information exchange between them. These principles include the locality principle stipulating that rules for synaptic weight adjustment can include only parameters of activity of the pre- and post-synaptic neurons. A particular case of this general principle is the well-known Hebbian plasticity rule in accordance with which the synapses frequently receiving spikes short time before postsynaptic spike generation are potentiated while the synapses with uncorrelated pre/post-synaptic activity are suppressed or are not modified.

The locality principle is very general and does not fix exact relationship between weight dynamics and pre/post-synaptic activity characteristics. For this reason, a plenty of very different SNN synaptic plasticity rules have been proposed, that makes the situation with SNN learning strikingly different from the uniform approach to learning in traditional neural networks. The majority of these rules can be considered as generalizations of spike timing dependent plasticity (STDP) – the synaptic plasticity rule, experimentally observed in living neurons [6]. The pure "classic" STDP can hardly be used as a basis of implementation of learning in SNN, especially in recurrent SNN, due to its inherent instability – in accordance with STDP, a potentiated synapse automatically gets more chances to be potentiated further causing the "runaway" network behavior. It made researchers to invent STDP generalizations [7 - 9] leading to more balanced network dynamics. Further neurophysiological studies also showed that synaptic plasticity in different kinds of neurons in different brain regions are described by various plasticity models sometimes greatly declining from the classic STDP [10].

This fact enables us to think that different plasticity models are adequate for different tasks. In fact, even in the realm of traditional neural networks, network architectures and synaptic weight tuning algorithms are different for, say, supervised and unsupervised learning. While the layered feed-forward networks are commonly used for supervised learning, for unsupervised learning, the flat networks with (implicit) lateral inhibition like Kohonen SOM [11] proofed their efficiency.

This thesis determined the motivation for this study. We would like to find SNN plasticity rules satisfying the locality principle and optimized for solution of one class of learning problem, namely, unsupervised learning. In the next Section, I formalize SNN unsupervised learning as a problem of finding spike frequency correlations. Further, I describe the novel synaptic plasticity model and the unsupervised learning algorithm SCoBUL based on it and show that it fits the specific requirements imposed by this representation of the unsupervised learning problem. In Section 4, a proof-of-concept level experimental evidence of SCoBUL efficiency using emulated DVS camera signal is presented.

## 2    Unsupervised learning in spiking domain

The problem of supervised learning in the most general case can be formulated as a search for certain features in the given dataset which distinguish it from the dataset with the identical statistical parameters (such as mean or standard deviation) but where each value is generated by the respective random number generator independently of other values. Nothing can be learnt from the data where each individual value is generated by independently working random number generators. Presence of some hidden structure, patterns which can be recognized by supervised learning algorithms is indicated, in the general case, by increased (or decreased) probability that certain values appear in certain places in the dataset comparatively to the situation when all data are completely random. These probability deviations can be expressed in terms of correlations (or anti-correlations) between certain variables included in the dataset or calculated as derivative variables. This view on unsupervised learning covers temporal patterns, as well – in the form of the correlations between a variable and the explicit time coordinate considered as an additional variable of the dataset or the correlations between the current value of a variable and the values of some variables (including this variable itself – in case of autocorrelations) in the past.

All this remains valid for data represented as spike sequences, but in this case the data values are extremely simple – they are in fact Boolean (spike / no spike). Thus, we can say that unsupervised learning problem for SNN can be formulated without loss of generality as a problem of detecting correlations between spike frequencies in input spike trains. Detecting of anti-correlations is also covered by this approach due to use of inhibitory neurons capable of implementing the logical operation NOT. If network includes the excitatory neuron A that would be constantly active unless it is not suppressed by the inhibitory neuron B then the spike sequence generated by the neuron A can be considered as a result of logical NOT applied to the input signal activating the neuron B. Then, the signals anti-correlated with B's input signal are correlated with A's output signal. Temporal (auto)correlations and correlations with time lag can be detected in the same way as for correlations between current spike frequencies if a network with some memory mechanism is used. For example, the well-known technique called liquid state machine (LSM) [12] utilizes this idea. LSM includes a large chaotic recurrent SNN whose role is to convert temporal dynamics of input signal into a very high dimensional representation in the form of current firing frequencies of its neurons. It is possible because reaction of SNN to an external stimulus manifested in terms of neuron firing frequencies may be observed during long time after the stimulus presentation.

It follows from the discussion above that, in case of SNN, the problem of unsupervised learning can be represented as a problem of detection of spike frequency correlations for spikes emitted by input nodes or neurons of the network. Let us note that time scales of these correlations may be different – from exact coincidence of firing times to cases of concurrent elevation of mean spike frequency measured for long time intervals.

It should be noted that such an understanding of unsupervised learning is very natural for spiking neurons. Indeed, the most basic operation characteristic for all spiking neuron models is detection of coinciding arrival of spikes to its synapses. Only when several synapses with sufficiently great weights receive spike inside more or less narrow time window, the neuron fires indicating this fact.

In order to define the solved problem formally, we consider the following simplified but still quite general input signal model. We assume that the input signal is generated by $N + 1$ Poissonian processes. One of them works always and plays the role of background noise. Let us denote its intensity as $p_0$. The other processes numbered by the index $i$ are switched on randomly with the probability $P_i$ and operate during the time interval $t_i$. During this time interval a certain set of input nodes (we will call it *cluster*) $\mathbf{C}_i$ emit spikes with the probability $p_0 + p_i$. This elevated activity of cluster's nodes will be called *pattern*. Evidently, the activity of input nodes inside every cluster $i$ is correlated and statistical significance of this correlation is determined by $p_i$ and the number of activations of this cluster in the whole observed input signal $n_i$. The goal of unsupervised learning is to teach SNN to react specifically to these patterns in the input spike stream. Namely, due to the appropriate synaptic plasticity rules, for each cluster, a recognizing neuron should appear in the network. This neuron should fire when and only when the respective cluster is active. Thus, our problem is parametrized by the value of $p_0$ and $N$ corteges <$\mathbf{C}_i$, $n_i$, $t_i$, $p_i$>.

## 3    The algorithm SCoBUL - network, neuron, synaptic plasticity.

In this work, I describe a novel SNN unsupervised learning algorithm approaching the unsupervised learning problem from the positions of spike frequency correlations. The algorithm is called SCoBUL (Spike Correlation Based Unsupervised Learning). An application of SNN to any problem related to learning includes three major logical components: network architecture, neuron model and plasticity rule. The novelty of the present work and the algorithm SCoBUL belongs mainly to the third component; while the network structure and the neuron model used here are quite common.

Similar to the majority of studies devoted to unsupervised learning, I utilize the so-called winner-takes-all (WTA) network architecture [13]. It is a one-layer SNN where every neuron is connected to a certain subset of input nodes (possibly, to all of them) thru excitatory links and has strong lateral inhibitory projections to the other neurons. This structure can be considered as a spiking analogue of Kohonen's self-organizing map [11], a very efficient architecture of traditional neural networks used in unsupervised learning tasks. The general idea of WTA is the following. Every neuron due to the respectively selected plasticity model tries to detect sets of input nodes with coinciding activity periods. At the same time, a neuron having learnt successfully to recognize such a group of correlated input nodes inhibits recognition of the same group by the other neurons blocking their activity by its inhibitory spikes emitted during activation of this group. Many extensions of this simple architecture have been proposed (for example, 2-layer WTA networks [14]) but, as it was said above, the direction of our movement is enhancement of the synaptic plasticity model.

Only one important novelty related to network structure is introduced in this work – the network structure is variable – the neurons may die and be born again (or migrate, if you like…). Neuron may die if it is constantly inhibited and cannot fire for a long time. In this case, it is destroyed and re-created by the same procedure which was used for construction of the original neuron population at the beginning of the simulation. Due to this feature, a neuron, inhibited by its happier neighbors having managed to recognize the most significant correlations in the input signal, gets a chance to be resurrected with a new combination of synaptic weights, which could help it to recognize some still "unoccupied" weakly correlated input node set

The neuron model utilized is also very simple, probably, the simplest spiking neuron model used in research and applications. It is leaky integrate-and-fire (LIF) neuron [15]. Its simplicity makes it efficiently implementable on the modern digital (TrueNorth [16], Loihi [17]) and even analog (BrainScaleS[18], NeuroGrid [19]) neurochips. I also use the simplest synapse model – current based delta-synapse. When such a synapse receives spike, it immediately increases (or decreases – if it is inhibitory) neuron's membrane potential by the value equal to its weight.

The SCoBUL synaptic plasticity model can be called a generalization of STDP but it is generalized and modified in several directions. Below, we will consider them and discuss how they help solve the unsupervised learning problem formulated in the previous section. Similar to the classic STDP, weight modifications in SCoBUL also depend on relative timing of pre- and post-synaptic spikes and this dependence includes a temporal parameter $\tau_P$ determining length of time interval inside which the pairs of spikes are considered as interrelated and can change synaptic weight. Describing the SCoBUL plasticity model below, I use the notion of *plasticity period*. Plasticity period is a time interval of length $2\tau_P$ centered at the moment of the postsynaptic spike but only if this postsynaptic spike is emitted after $\tau_P$ or more time since the center of the previous plasticity period. It is important to note also that inhibitory connections are not plastic in this model.

### 3.1 Synaptic resource

The classic form of STDP has additive character. In accordance with STDP, synaptic weight is increased or decreased by a certain value depending on relative position of the pre- and post-synaptic spikes on the time axis. If this rule is applied without any restrictions or corrections then it can easily lead to senseless very high positive (or negative) weights due to STDP's inherent positive feedback. To prevent this, the values of synaptic weights are bounded artificially by a certain value from above and by zero from below. It solves the problem of unlimited synaptic weights but causes another problem – of catastrophic forgetting. Indeed, let us imagine that network was being trained to recognize something for a long time. As a result, the majority of synaptic weights became either saturated (equal to the maximum possible value) or suppressed (equal to 0). However, presentation of even few wrong training examples or examples containing other patters or simply noise is sufficient to destroy the weight configuration learnt and nothing can prevent it. The network will forget everything it has learnt. In order to fight this problem, it was proposed in my several earlier works [20] to apply additive plasticity rules to the so-called *synaptic*

*resource* instead of the synaptic weight. The value of synaptic resource $W$ depends monotonously on the synaptic weight $w$ in accordance with the formula

$$w = w_{\min} + \frac{(w_{\max} - w_{\min})\max(W,0)}{w_{\max} - w_{\min} + \max(W,0)}. \qquad (1)$$

In this model, the weight values lay inside the range [$w_{min}$, $w_{max}$) - while $W$ runs from $-\infty$ to $+\infty$, $w$ runs from $w_{\min}$ to $w_{\max}$. When $W$ is either negative or highly positive, synaptic plasticity does not affect a synapse's strength. Instead, it affects its *stability* – how many times the synapse should be potentiated or depressed to move it from the saturated state. Thus, to destroy the trained network state, it is necessary to present the number of "bad" examples close to the number of "good" examples used to train it. It should be noted that this feature was found to be useful not only for unsupervised learning – we use it in all our SNN studies. Let us add that in the present research $w_{\min}$ is set equal to 0 everywhere.

## 3.2 Unconditional synapse depression

When a synapse receives spike, its synaptic resource is decreased by the constant value $d$ but this decrease can happen at most once inside any time window of length $2\tau_P$. Why this simple rule is useful, we will see later – when other features of the SCoBUL model will be discussed.

## 3.3 Constant symmetric STDP

In my model, all presynaptic spikes arriving inside a plasticity period strengthen the synapse. However, a synapse can be potentiated at most once inside one plasticity period – by the spike coming first. It should be stressed that the relative order of pre- and post-synaptic spikes is not important. When presynaptic spike comes just after postsynaptic spike, it potentiates the synapse as well. Thus, this rule can be called symmetric STDP. Besides that, the value of the synaptic resource increment is the constant $D^+$, it does not depend on the exact time difference between pre- and post-synaptic spikes.

## 3.4 Suppression of strong inactive synapses

This rule is a conceptually new addition to the classic STDP. It states that if a synapse with positive resource has not received a spike during current plasticity period it is depressed at its end by the constant $D^-$.

## 3.5 Constant total synaptic resource

The last important logical component of SCoBUL is constancy of neuron's total synaptic resource. Every time some synapse is potentiated or depressed, the resources of all other synapses are modified in the opposite direction and by the same value calculated from the condition that the total synaptic resource of the neuron should remain the same.

Now, having described all logical components of the SCoBUL plasticity model, let us analyze and explain them from point of view of unsupervised learning problem formulated at the end of the previous section. Let us begin from point 3.2. In conjunction with point 3.5, it gives the following very useful effect. The classic STDP has many drawbacks, and one of them is uselessness of silent neurons. Indeed, in the classic STDP model, the process of weight modification is bound to firing. The neurons which do not fire are not plastic. Therefore, if some neuron is silent because it is constantly inhibited by other neurons, it will stay in this state forever, and, therefore, will be just useless burden consuming computational resources but producing nothing. In my model, combination of rules 3.2 and 3.5 gives the following effect. Activity of certain sets of input nodes makes to fire some neurons. Inhibition from these active neurons forces the synapses of silent inhibited neurons connected to the active input nodes redistribute their synaptic resource to the other synapses, connected with less active input node groups. Even if these weak input node groups could not force to fire any neuron in the initial network configuration, after this resource redistribution, some silent neurons may accumulate in the respective synapses the amount of synaptic resource sufficient to fire. Thus, this process of "squeezing" synaptic resource out of active synapses to less active synapses helps the network recognize all correlations in the input spike streams – not only the most significant ones.

The fact that symmetric variant of STDP is more suitable for unsupervised learning than its classic asymmetric form is obvious. Indeed, activity period of correlated input node groups may be long. In case of possible random delays of spikes inside this activity time window, some of them may appear earlier, some – later. When neuron learns to recognize this group, its synapses connected to these nodes are strengthened. Therefore, it begins to fire earlier when this input node group gets activated. But it means that more of its synapses connected to these nodes will experience depression instead of facilitation. Then, further recognition improvement will be impossible. It is evident, that the symmetric STDP does not face this difficulty.

At last, point 3.4 solves another hard problem of unsupervised learning. It would be desirable that one neuron recognized one cluster, and one cluster were recognized by one neuron. The situation when two neurons recognize the same cluster is cured by introduction of stronger mutual inhibition – if the lateral inhibition is sufficiently strong, the state when several neurons recognize the same pattern is evidently unstable. The problem of recognition of several clusters by one neuron is much harder. Strong lateral inhibition cannot help here – instead it can make this undesirable state more probable. Other rules considered above cannot help as well. The rule 3.4 was designed especially to fight this unpleasant scenario. Indeed, if a neuron recognizes the clusters A and B, it means that its synapses connected to A and B are strong. Assume that A is active. The neuron fires and therefore, accordingly to rule 3.3, the connections leading to A are potentiated. However, the synapses connected to B have not received spikes during the respective plasticity period and are suppressed by rule 3.4. Thus, rule 3.4 makes the state when one neuron recognizes several independent correlated input node groups unstable.

This discussion demonstrated how rules 3.1 – 3.5 help address various aspects and complications of the general problem of unsupervised learning, namely:

- to recognize strong and weak correlations by the same network (to prevent situations when there are no recognizing neurons for some clusters);
- to make recognizing neurons sufficiently specific (to prevent situations when some clusters are recognized by several neurons);
- to make recognition unambiguous (to prevent situations when several clusters are recognized by the same neuron).

In the next section, we will consider the experimental confirmation of these theses on artificially generated data imitating a signal from DVS camera.

# 4    Experimental comparison of STDP and SCoBUL plasticity rules on an imitated DVS signal.

In order to evaluate the benefits of the SCoBUL plasticity comparatively to the standard STDP I decided to select a task close to real application of SNN, namely, processing of spiking video signal sent from a DVS camera. For this purpose, a program emulator of DVS camera has been created. To simplify and speed up the simulation experiments, I emulated small camera view field of size 20×20 pixels. 3 spike streams (channels) correspond to each pixel. Spike frequency in the first channel is proportional to the pixel brightness. The other two channels send spikes every time the pixel brightness increases or decreases by a certain value. Thus, the whole input signal includes 1200 spiking channels. The thresholds used to convert brightness and brightness changes into spikes are selected so that the mean spike frequency in all channels is close to 30Hz.

In these tests, I selected a very simple picture – a light spot moving in the view field of this imaginary DVS camera in various directions and with various speed. At every moment, coordinates and speed of light spot (i.e. the point in the 4-dimensional phase space occupied by it) are known. The task is to determine them from the current activity of the WTA network neurons.

More precisely, the procedure is the following. The whole emulation takes 3000000 time steps (we assume that 1 time step = 1 msec). The time necessary for the light step to cross the DVS view field is several hundred milliseconds. During first 2000 sec the network is trained. Next 600 sec are used to determine the centers of receptive field of every neuron in the phase space. Last 400 sec are broken to 40 msec intervals. For each interval, the real central position of the light spot in the phase space and the predicted value of this position are determined. The predicted position is the weighted mean of neuron receptive field centers where the weight is amount of spikes emitted by the given neuron in this time interval. The value of mean squared distance between the real and predicted light spot positions in each time interval serves as a measure of unsupervised learning success. Small value of this difference would be an evidence that network's neurons learnt to recognize specific positions of the light spot in the phase space. Euclidean metrics in the phase space was selected so that the standard deviation of light spot coordinate during the whole simulation period would be the same for all coordinates.

I performed this test with the networks with the standard STDP and the SCoBUL plasticity. In order to make this competition fair I used in both cases the same network parameter optimization procedure based on genetic algorithm. The parameter variation ranges were the same or equivalent for both plasticity models; I also made sure that the optimum parameter values found were not close to the boundaries of the search space. To diminish the probability of accidental bad or good result, I took the criterion values averaged for 3 tests with the same hyperparameters but with different sets of initial synaptic weight values. The population size was 300, mutation probability per individual equaled to 0.5, elitism level was 0.1. Genetic algorithm terminated when new generation did not show criterion improvement. The optimized (minimized) parameter was the mean squared distance between the real and predicted light spot position divided by the mean squared distance between spot position and the centrum of all spot positions during the entire emulation. It is called "Normalized mean squared distance" on Figure 1 showing the results obtained by genetic algorithm.

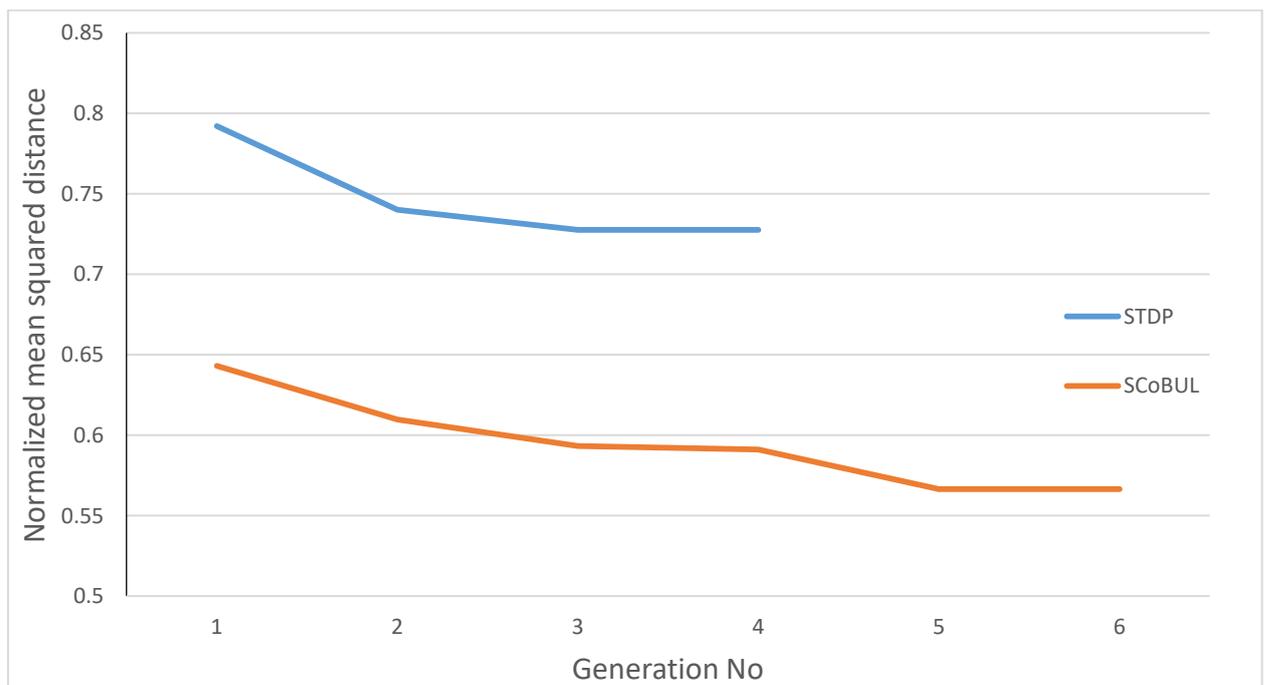

Figure 1. The course of minimization of the light spot position determination error in sequential generations of genetic algorithm for STDP and SCoBUL synaptic plasticity rules.

We see that 6 generations were required for SCoBUL networks to reach criterion value stabilization, while in case STDP the stabilization was reached earlier – after 4 generations; and we see that SCoBUL networks give much more accurate light spot position/speed determination than STDP networks. While these results should be considered as preliminary and they should be verified in other unsupervised learning tasks, the supremacy of SCoBUL over the classic STDP in this case is undoubted.

## 5    Conclusion.

In this paper, the problem of unsupervised learning of SNN was re-formulated as a problem of finding spike frequency correlations in input spike streams. Using this approach and remaining inside the boundaries of the synaptic plasticity rule locality principle, I propose a modification of the classic STDP model, which optimizes it for unsupervised learning. Since our research project is oriented primarily to application of SNN to processing of sensory data represented in the spiking form and, as its most important particular case, to processing of DVS-generated signal, I used an artificially generated "DVS signal" as a benchmark to compare the standard STDP-based WTA network and the SCoBUL network. It was found that the SCoBUL model gives significantly better results. This result can be evaluated as promising and opening the way to further perfection of our model while more exact and complete measurements of its properties and possible limitations are obviously needed.

The other goal of this research is creation of hardware-friendly version of STDP-like synaptic plasticity model. In SCoBUL, this goal is achieved due to the special scenario of application of rules 3.1 – 3.5. Some of these rules (3.2 – 3.4) are bound to pre- and post-synaptic spikes and therefore are applied very often. However, these rules are very simple – they have the form of addition or subtraction of certain constant values. Rules 3.1 and 3.5 (the synaptic resource renormalization and the calculation of synaptic weight from synaptic resource) include much more expensive operations like multiplication and division. However, it is admissible to apply them periodically with sufficiently long period (say, once per second). Thus, in general, SCoBUL may have more economic and/or fast implementation than the standard STDP – it depends on the concrete processor architecture used.

## Acknowledgements.

The present work is a part of the research project in the field of SNN carried out by Chuvash State University in cooperation with Kaspersky and the private company Cifrum.

Preliminary computations resulting in creation of the SCoBUL algorithm were performed on the computers belonging to Kaspersky, Cifrum and me. Cifrum's GPU cluster was used for running the optimization procedure reported in Section 4.

## References.